\documentclass[letterpaper, 10 pt, conference]{ieeeconf}  

\IEEEoverridecommandlockouts                              

\usepackage{graphics} 
\usepackage{epsfig} 
\usepackage{amsmath} 
\usepackage{amssymb}  
\usepackage{subfigure}

\DeclareMathOperator*{\argmin}{argmin}

\title{\LARGE \bf
Learning to Autonomously Reach Objects with NICO and Grow-When-Required Networks
}

\author{Nima Rahrakhshan$^{1}$, Matthias Kerzel$^{2}$, Philipp Allgeuer$^{2}$, Nicolas Duczek$^{3}$, and Stefan Wermter$^{2}$
\thanks{*This research was partially supported by the Federal Ministry for Economic Affairs and Climate Action (BMWK) under the project KI-SIGS and the German Research Foundation (DFG) under the project Transregio Crossmodal Learning (TRR 169). 
}
\thanks{$^{1}${\tt\small nrahrakhshan@icloud.com}}%
\thanks{$^{2}$Knowledge Technology, Department of Informatics, University of Hamburg, Germany 
        {\tt\small \{kerzel, allgeuer, wermter\} @ informatik.uni-hamburg.de}}%
\thanks{$^{3}${\tt\small nduczek@icloud.com}}}

\begin{document}

\maketitle
\thispagestyle{empty}
\pagestyle{empty}

\begin{abstract}
The act of reaching for an object is a fundamental yet complex skill for a robotic agent, requiring a high degree of visuomotor control and coordination. In consideration of dynamic environments, a robot capable of autonomously adapting to novel situations is desired. In this paper, a developmental robotics approach is used to autonomously learn visuomotor coordination on the NICO (Neuro-Inspired COmpanion) platform, for the task of object reaching. The robot interacts with its environment and learns associations between motor commands and temporally correlated sensory perceptions based on Hebbian learning. Multiple Grow-When-Required (GWR) networks are used to learn increasingly more complex motoric behaviors, by first learning how to direct the gaze towards a visual stimulus, followed by learning motor control of the arm, and finally learning how to reach for an object using eye-hand coordination. We demonstrate that the model is able to deal with an unforeseen mechanical change in the NICO's body, showing the adaptability of the proposed approach. In evaluations of our approach, we show that the humanoid robot NICO is able to reach objects with a 76\% success rate.
\end{abstract}

\section{Introduction}

To reach for an object, a robotic agent must first locate the object in its surrounding space, and then move its hand towards it. The robotic agent needs to process the raw sensory data, while at the same time coordinating the motors, requiring a high degree of visuomotor control. Furthermore, dynamic elements in the environment may introduce temporary or permanent changes in the surrounds or in the robotic agent's bodily characteristics, which can strongly affect the perceived information of the agent \cite{schillaci_exploration_2016}.
Therefore, a robot must be able to continuously learn from its interactions with the environment, to react and adapt to unexpected events \cite{schillaci_sensorimotor_2014}.

\begin{figure}
    \centering
    \includegraphics[width=1.0\columnwidth]{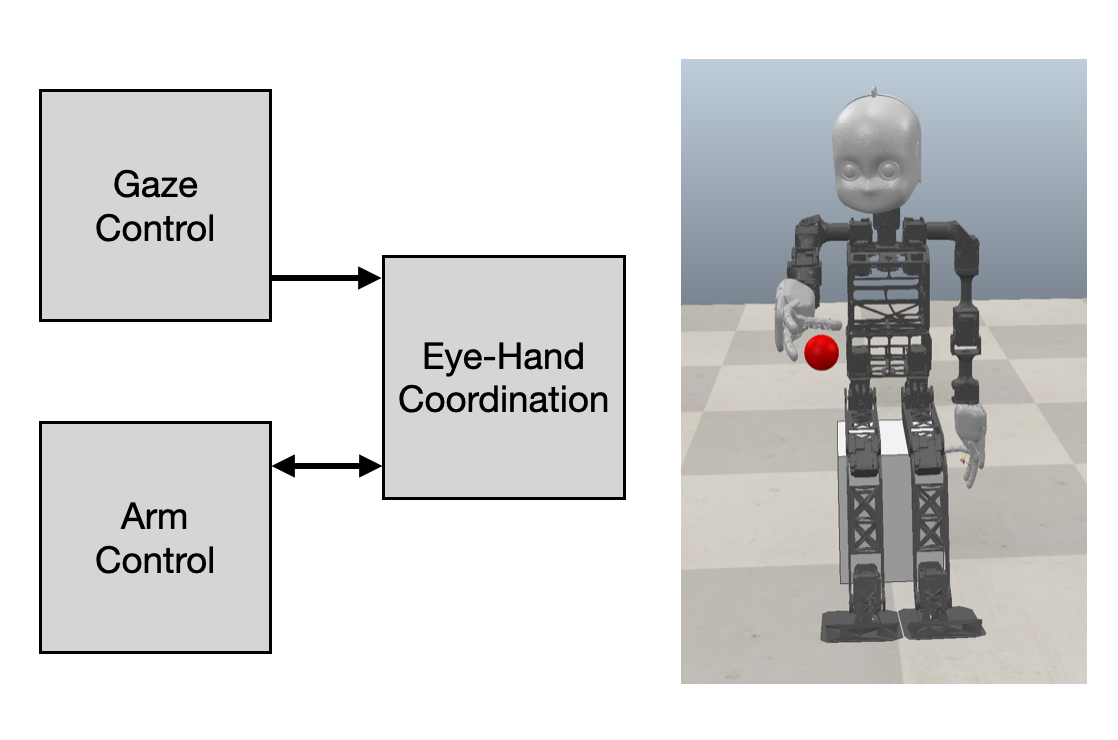}
    \caption{Left: Training curriculum for our embodied neural models. Right: Experimental setup with the humanoid robot NICO, reaching the target within the reachable space.}
    \label{fig:teaser_image}
    \vspace{-12pt}
\end{figure}

To address this issue, we propose a neural model for robotic visuomotor learning based on a growing self-organizing model \cite{marsland_self-organising_2002}, that continually learns the relationship between the visual and motoric information streams acquired through the body-environment interaction. We train and evaluate the approach on the humanoid robot NICO, the Neuro-Inspired COmpanion \cite{kerzel_nico_2017, kerzel_nico_2020} in a virtual environment, depicted in Figure \ref{fig:teaser_image}. We show the adaptability of the approach, and investigate if the robot is able to autonomously learn to reach objects. We investigate how the internal model copes with mechanical or environmental changes over time.

Our approach is based on the developmental robotics research paradigm, which investigates behavioral components established from developmental sciences \cite{lungarella_developmental_2003} to design computational models that can be embedded into robotic agents. A central idea of developmental robotics is the embodiment, such that the robotic agent has a body that mediates perception, affects behavior, and shapes the way it interacts with its surrounding environment \cite{lungarella_developmental_2003,lara_embodied_2018}.
To learn visuomotor control, our computational model is embedded into a humanoid robotic agent---it learns based on the sensorimotor information generated from the agent's bodily interaction with the external environment.

Our model first learns individual unimodal information maps for each of the various sensory and motor data streams, then learns associations to create multimodal connections with Hebbian learning between the different maps. The learning stage consists of three parts for the acquisition of reaching skills with the humanoid robot NICO, which are developed autonomously initially through random motor babbling. First, gaze control is learned to situate NICO in the surrounding environment, then arm control is learned, and in the final learning step, transfer learning from gaze and arm control is used to develop eye-hand coordination.

In summary, this paper contributes a novel continual learning approach for a robotic reach-for-grasp task that is based on highly adaptable self-organizing Grow-When-Required networks. In our experiments, the approach enables NICO not only to learn to reach for objects with a 76\% accuracy, but also to autonomously adjust to motor errors in a continuous learning scenario.

\section{Related Work}

Visuomotor learning emerges from long-term interactions with the surrounding environment. Through this body-environment interaction, the robotic agent gains multimodal information from its various sensors, which have been proposed to be integrated into a body schema \cite{hoffmann_body_2010}. The body schema allows integrating information from various sensory and motor streams to keep an up-to-date representation of the positions of the different body parts in space. These body schemas represent an internal model for the robotic agent, which can be fundamental for internal simulation processes and bridge the gap between low-level sensorimotor representations and basic cognitive skills \cite{schillaci_exploration_2016, nguyen_2021}. 

Inspired by the studies on body representations suggesting the existence of topographic maps in the brain \cite{udin_formation_1988,cang_developmental_2013}, which have been reported to be fundamental for sensorimotor processing and learning \cite{kaas_topographic_1997}, interest in developing self-organizing computational models 
has been growing \cite{schillaci_exploration_2016}. These multimodal body representations self-organize and adapt over the sensorimotor experience acquired through the body-environment interaction.

Throughout the recent research on the autonomous acquisition of reaching in humanoid robots, self-organizing maps (SOM) \cite{kohonen_self-organizing_1990} or different variants of SOMs have been in the focus of research---see Schillaci et al. \cite{schillaci_exploration_2016} for an overview. Essentially all the listed approaches have in common that the network structure needs to be defined a priori, which makes them unsuitable for dynamic environments or continuous learning tasks \cite{marsland_self-organising_2002}. Visuomotor learning should ideally be considered as a continuous learning task, in which the internal model adds new neurons whenever the model cannot sufficiently match the present sensorimotor experiences.

Our experimental design and training curriculum for our embodied neural models is inspired by human infant development: Infants show a remarkable talent for developing visuomotor control and coordination at early stages, as well as rapid cognitive growth \cite{law_infant_2011}. 
As summarized by Law et al. \cite{law_infant_2011,law_saccades_2014,law_psychology_2014}, we find similar main stages of visuomotor control and coordination for robotic agents, i.e, gaze control, arm control and eye-hand coordination. 
Gaze control is the ability to locate a salient region, or region of interest, in the perceived scene and then direct the visual system towards it, such that the salient region is centered. The robotic agent becomes situated, as it then has the ability to react and interact with its surrounding environment \cite{hendriks-jansen_catching_1996}. Arm control, on the other hand, is the ability to move the hand towards a location within the reachable space. The robotic agent needs to be able to move the hand towards a point in space at which a target is located. Eye-hand coordination is the coordinated movement of the gaze and the hand, which is a required motor skill for pointing and/or reaching behaviors \cite{lungarella_beyond_2003}. The gaze and the hand should be able to move freely, 
but should be associated with each other such that eye-hand correlation and coordination can be supported \cite{law_psychology_2014}. The hand's location needs to be associated with a corresponding direction of gaze, which is accomplished by first moving the hand towards a location in space and then directing the gaze towards the hand. The reach action is the exact inverse action of the training process \cite{shaw_representations_2015}, that is, to first direct the gaze towards a visual stimulus in the perceived scene, e.g., an object, and then move the hand towards it. 

\section{Visuomotor Learning}

\begin{figure*}
    \centering \includegraphics[width=\textwidth]{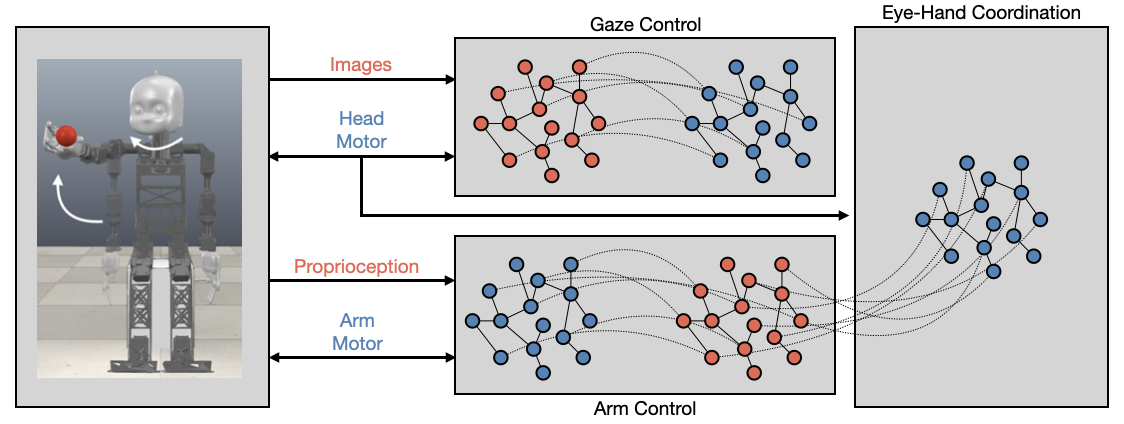}
    \caption{The proposed system for the autonomous acquisition of reaching skills in the humanoid robot NICO. First, our neural model uses Grow-When-Required networks to train unimodal information separately and then creates multimodal connections between relevant information.
    For the gaze control learning stage, our model creates multimodal associations between the head motor commands and the perceived sensory change in the form of images. For the arm control learning stage, our model creates multimodal associations between the arm motor commands and the proprioceptive information of NICO's hand in space. The final learning stage, eye-hand coordination, links both learning stages by using transfer learning and creating multimodal associations between the absolute head motor values and the proprioceptive information.}
    \label{fig:overview}
    \vspace{-12pt}
\end{figure*}

\subsection{The Grow-When-Required Network}

The Grow-When-Required (GWR) network by Marsland et al. \cite{marsland_self-organising_2002} is a sheet-like self-organizing growing neural network, which adds neurons to the network whenever it cannot sufficiently match the input. The GWR has a large collection of neurons, each with their associated weight vectors and lateral connections, which connect neurons that represent similar perceptions. Given a learning input $x(t)$ presented to the network $S$, the best matching unit (BMU) $b$ and second BMU $s$ are computed as follows:
\begin{align}
    b = \underset{i \: \in \: S}{\argmin} { \: || x(t) - w_i || }, \\
    s = \underset{i \: \in \: S \setminus \{b\}}{\argmin} { \: || x(t) - w_i || },
\end{align}
where a connection $(b, s)$ is created if there is none yet, otherwise the age of the connection is set to zero. 
Each neuron is equipped with a habituation counter $h_i \in [0,1]$, counting how frequently a neuron has been selected as the BMU. The habituation at a time step for the BMU $b$ and its neighbors $n$ is calculated as
\begin{align}
    h_{i}(t + 1) = \tau_i * 1.05 * (1 - h_{i}(t)) - \tau_i,
\end{align}
with $i \in \{b, n\}$, where $\tau_i$ is a constant that regulates a decreasing response and $\tau_n < \tau_b$. 
The activity of the BMU $b$ is computed as
\begin{align}
    a_{b} = e^{- || x(t) - w_b || },
\end{align}
where $w_b$ is the weight vector associated with the BMU $b$. Given a predefined activity threshold $a_T$ and habituation threshold $h_T$, a new neuron is added into the network if the activity 
is below the activity threshold $a_{b} < a_T$ and the habituation counter is below the habituation threshold $h_{b} < h_T$. The activity threshold $a_T$ and the habituation threshold $h_T$ regulate how many neurons and how fast neurons are added to the network. 
The weights of 
the BMU $b$ and all its neighbors $n$ are updated using
\begin{align}
    w_b = \epsilon_b * h_{b} * (x(t) - w_b), \\
    w_n = \epsilon_n * h_{n} * (x(t) - w_n),
\end{align}
where $\epsilon_b$ and $\epsilon_n$ are fixed learning parameters such that ${0 < \epsilon_n < \epsilon_b < 1}$. Lastly, the algorithm removes 
all connections with an age larger than the maximum allowed age and removes all isolated neurons. These steps are repeated for a set number 
of epochs, or until a stopping criterion is reached.

\subsection{Neural Model for Visuomotor Learning}

As a robot interacts with its environment, highly correlated motor $M$ and sensory $S$ data streams are perceived. For our proposed model, we start by training two GWRs, one for the motor modalities and one for the sensor modalities. The model hyperparameters are summarized in Table \ref{tab:model_parameter}. Once the GWRs are trained, multimodal associations in the form of neural connections are made between the two GWRs by iterating through the sensorimotor training samples and searching for the pairs of best matching units in each map:
\begin{align}
      s = \underset{i}{\argmin} { \: || S - w_i || } \\
      m = \underset{j}{\argmin} { \: || M - w_j || }
\end{align}
After the BMU is found in both maps, a direct connection is built via a positive Hebbian learning rule based on the activities $a_s$, $a_m$ of the two BMUs in their respective map:
\begin{align}
      w_{sm}(t + 1) = w_{sm}(t) + \alpha * a_{s} * a_{m}, \label{eq:association}
\end{align}
where $w_{sm}$ is a direct weight connecting neuron $s$ in the sensory map and neuron $m$ in the motor map.
The Hebbian learning parameter $\alpha$ is set to $0.5$ throughout all the experiments. 
Once the system has fully iterated through the sensorimotor training samples and generated the required multimodal associations, 
the model can be queried by finding the sensory BMU that best matches an observed $S$, and determining which neuron in the motor map is co-activated most strongly. This can then be mapped to a corresponding motor command.

Overall, the proposed neural architecture consists of three submodules for visuomotor learning, depicted in Figure~\ref{fig:overview}. 
Gaze control is learned first to situate a robotic agent in the surrounding environment, then arm control is learned, and finally, both modules are combined to learn eye-hand coordination.

\begin{table}[!t]
    \renewcommand{\arraystretch}{1.3}
    \caption{Training parameters for each GWR.}
    \label{tab:model_parameter}
    \centering
    \begin{tabular}{| l  l |}
        \hline
        Hyperparameter & Value \\ \hline
        Epochs & 40 \\ 
        Maximum Age & 5 \\ 
        Maximum Neurons & 6000 \\ 
        Learning Rate & $\epsilon_{b} = 0.5$, $\epsilon_{n} = 0.01$ \\ 
        Habiutation Rates & $\tau_{b} = 0.3$, $\tau_{n} = 0.1$ \\ 
        Activity Threshold & $a_{T} \in (0.0, 0.9]$ \\ 
        Habituation Threshold & $h_{T} \in (0.0, 0.9]$ \\ \hline 
    \end{tabular}
    \vspace{-12pt}
\end{table}

\section{Dataset Generation}

\subsection{The Humanoid Robot NICO}

The experiments are realized with a virtual NICO humanoid robot \cite{kerzel_nico_2017}, a developmental robot platform with rich sensory and motor capabilities, enabling embodied neuro-cognitive models. To enable human-like manipulation, each of NICO's arms have 6 degrees of freedom (DoF), in which 3 DoF are for the shoulder, 1 DoF for the movement of the elbow, and 2 DoF for wrist rotation and flexion. Furthermore, NICO's neck has 2 DoF to control the head's yaw and pitch, i.e. horizontal and vertical movements. The visual modality is realized in the form of two parallel cameras mounted in NICO's head. 
NICO has a further modality to provide a proprioceptive sense for all DoF, which is the information about motor values, movements, position, and forces.

\subsection{Environmental Setup}

For the experiment, a virtual, simplified realization of the humanoid robot NICO is provided 
in the robot simulation framework CoppeliaSim \cite{rohmer_coppeliasim} via the NICO API. The humanoid robot NICO is seated on a $25$ cm long cuboid in the virtual environment. The target to be reached is implemented as a red ball of $5$ cm in diameter placed in front of NICO, such that it is salient from the surrounding environment. As the experiment evaluates the capability of the proposed internal model for the task of reaching, only a single target is used. To reduce the dimensionality of the data and, therefore, the complexity of the experiments, only the camera 
in the right eye is used for the visual modality. Furthermore, only the right arm with four joints is used to evaluate the task of reaching, as the two joints that rotate the wrist do not influence the arm's position in space.

\subsection{Data Collection}

To collect the data and, therefore, the necessary information for visuomotor coordination, we adopt a random motor babbling strategy. For each learning stage, the following strategies were designed:

\emph{1) Gaze dataset:}
The target is initially placed in the center of the visual system of NICO, and random motor 
babbling is performed by randomly selecting a motor command $M_{\Delta head}$ while observing the change in the sensory data. The sensory data $S_{head}$ yields the centroid of the detected target in the image plane by first downscaling the image by a factor of $8$ to $80 \times 60 \times 3$ to reduce the dimensionality of the sensory data \cite{kerzel_visuomotor_2017}, and then applying a 
color thresholding algorithm to highlight the target. The sensorimotor experience is described by the pair $(S_{head}, -M_{\Delta head})$, which combines the current centroid $S_{head}$ with the additive inverse of $M_{\Delta head}$, as the exact opposite values of the current motor values are needed to bring the target again into the center of the visual system. We record a dataset over $1000$ iterations of this process, while also augmenting the observed data via interpolation, yielding $6410$ samples of sensorimotor experience.

\emph{2) Arm dataset:}
To collect the dataset for arm control, random motor babbling is performed on the joints of NICO's right arm. The sensorimotor experience is described by the pair $(S_{arm}, M_{arm})$, where $S_{arm}$ is the hand's current position in Cartesian space relative to a fixed point in the NICO's torso, and $M_{arm}$ are the current motor joint values. These datapoints are $(x, y, z)$-coordinates nominally given in centimeters for convenience. We record the dataset for $1000$ iterations, while also interpolating the data, yielding $14910$ samples of sensorimotor experience.

\emph{3) Eye-hand dataset:}
The dataset for eye-hand coordination is recorded by randomly moving the arm within the reachable workspace. In each iteration, a random motor command $M_{arm}$ is executed, and the internal model for gaze control is subsequently used to bring the hand into the center of the visual system. The hand is visually located by placing the red ball in the open grasp of the hand, i.e. such that the hand would grasp the ball if it were closed. If the hand moves out of the visual system, such that the hand cannot be found by the gaze control model, a simple algorithmic behavior moves NICO's head around until a salient region is found, or the iteration is canceled, e.g. in order to deal with the situation that the hand moved out of the gaze space.
The sensorimotor experience is described by a triplet of the current Cartesian location of the hand $S_{arm}$,
the current motor values for the joints of the right arm $M_{arm}$ and the current motor values for the joints of the head $M_{head}$. The dataset is recorded for 1000 iterations, yielding as many triplets of sensorimotor experience.

\emph{4) Environmental change:}
To test the proposed model's capability to react to environmental changes, a mechanical change in NICO's body is simulated in the form of a damaged joint, namely the upper joint in the right arm responsible for the arm's inward and outward rotation. The joint is fixed in its zero position, which alters the reachable space. The dataset recording was conducted in the same manner as in the eye-hand coordination experiment. However, the dataset collection only lasted for $500$ iterations, which is half the number used for the eye-hand coordination training. 

\section{Experimental Results and Discussion}

\subsection{Gaze Control}

The hyperparameter optimization led to values of ${a_T = 0.5}$, ${h_T = 0.7}$ with $3306$ neurons in total in the sensory map, and values of $a_T = 0.9$, $h_T = 0.3$ with $6000$ neurons in total in the motor map. In the sensory map, $3014$ neurons built a cross-modal connection, while $292$ have no connection to the motor map.
In the motor map, $4633$ neurons built a connection, while $1367$ neurons have no connection to the sensory map. 
The model is tested for $1000$ iterations for the ability to control the gaze with the NICO in the simulation environment, repeated $5$ times. The median Euclidean error of the gaze centering is $2.2(\pm 0.0)$ pixels, an overview of the results is in Table \ref{tab:results}. 

The results for gaze control are limited by the resolution of the used images, and higher resolution images would allow for more fine-grained head motor control, but also increase the complexity of our model. However, our model successfully learns to control the gaze and move the head to center the target in the visual field, and is therefore well suited for the reaching task.

\subsection{Arm Control}

\begin{figure*}
    \centering 
    \subfigure[Arm control learning stage.]{\includegraphics[width=1.0\columnwidth]{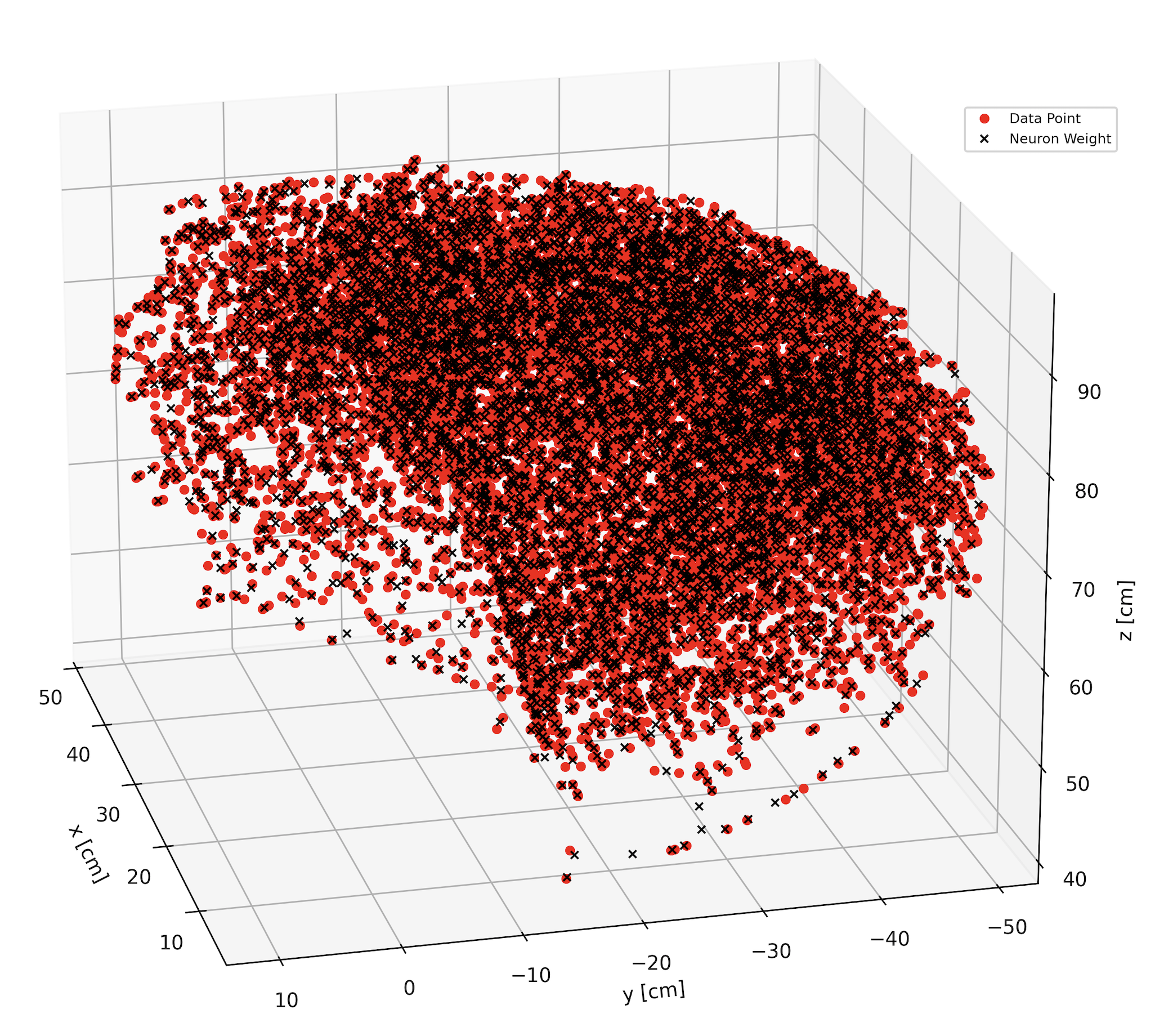}}
    \subfigure[Eye-hand coordination learning stage.]{\includegraphics[width=1.0\columnwidth]{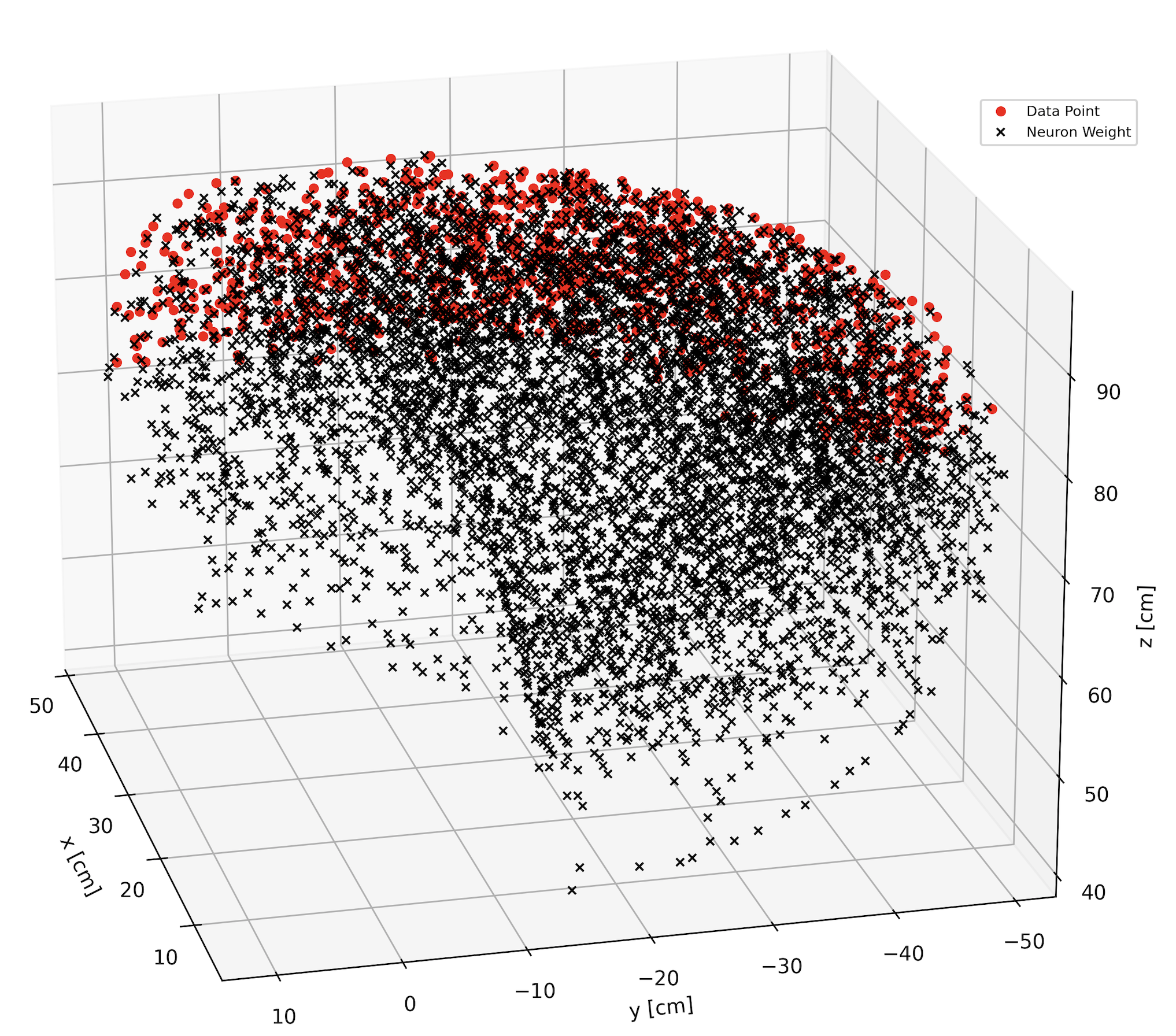}}
    \caption{The weights of our neural model and sensorimotor experience during arm control and eye-hand coordination.}
    \label{fig:arm_resuls}
    \vspace{-12pt}
\end{figure*}

For the arm control model, the hyperparameter optimization leads to $a_T = 0.5$, $h_T = 0.7$ with $6000$ neurons in the sensory map, and $a_T = 0.1$, $h_T = 0.5$ with $6000$ neurons in the motor map. In the sensory GWR, $299$ neurons have no connection, while $119$ neurons have no connection in the motor GWR. The sensorimotor experience and the weights of the neural model are depicted in Figure \ref{fig:arm_resuls}a.
The internal model is tested for $1000$ iterations for the ability to control the arm with the humanoid NICO in the simulation environment, repeated $5$ times. The median Euclidean error is $1.10(\pm 0.03)$ cm, the lowest error is $0.50(\pm 0.01)$ cm, and the highest error is $25.96(\pm 1.95)$ cm. The error is mainly influenced by the maximum number of neurons, which was capped at $6000$, which was set to prevent overfitting and encourage the network to generalize. Our model learned to control the arm and move the hand towards the desired location in the surrounding space. The median errors are deemed to be within tolerance for the reaching task.

\subsection{Eye-Hand Coordination}

\begin{table}[!t]
    \renewcommand{\arraystretch}{1.3}
    \caption{The error for the different learning stages.}
    \label{tab:results}
    \centering
    \begin{tabular}{| l | c | c | c |} \hline
        Learning stage & Median error & Min error & Max error \\ \hline
        Gaze (px) & $2.2 \: (\pm 0.0)$ & $0 \: (\pm 0.0)$ & $51.4 \: (\pm 0.0)$ \\
        Arm (cm) & $1.1 \: (\pm 0.03)$ & $0.50 \: (\pm 0.01)$ & $25.96 \: (\pm 1.95)$ \\
        Eye-Hand (cm) & $2.44 \: (\pm 0.1)$ & $1.38 \: (\pm 0.61)$ & $16.63 \: (\pm 8.01)$ \\
        Env. Change (cm) & $5.08 \: (\pm 1.7)$ & $2.76 \: (\pm 2.76)$ & $26.70 \: (\pm 8.84)$\\ \hline
    \end{tabular}
    \vspace{-12pt}
\end{table}

The hyperparameter optimization for the GWR trained with the absolute head motor values leads to values of $a_T = 0.5$, $h_T = 0.9$, and a network with $1000$ neurons. Transfer learning is used on the GWRs of the arm control model with the acquired sensorimotor experience for eye-hand coordination. The sensorimotor experience and the weights of the neural model for eye-hand coordination are depicted in Figure \ref{fig:arm_resuls}b.
The number of neurons in the arm control model decreases to $5925$ and $5937$ neurons respectively for the sensory and motor maps. The model for eye-hand coordination is then built between the motor map trained on the absolute head motor values, and the sensory map trained on the Cartesian hand position values, which results in $950$ neurons in the absolute head motor map, and $791$ neurons in the Cartesian map having built a connection, respectively.

The model is tested for $1000$ iterations, repeated $5$ times, for the ability to reach the target with the humanoid robot NICO. 
The median Euclidean error from the gaze control model is $2.23(\pm 0.0)$ pixels, while 
the median Euclidean error for the arm control model is $2.44(\pm 0.10)$ cm. The minimum error is $0.68(\pm 0.95)$ cm, while the maximum error 
is $16.63(\pm 8.01)$ cm. The highest errors stem from NICO reaching too far or too short along the depth axis, which is in part due to motoric redundancy between the visually perceived target and arm motor values. This redundancy stems from the unavailability of
depth information for a target, as multiple different positions on the depth axis result in the same visually perceived ball location, and therefore also in the same absolute head motor values. 
The target is successfully reached $762 (\pm 6.72)$ times, judged by the condition of graspability with a tolerance of $\pm 3$ cm. Overall, this corresponds to a $76.2(\pm 0.67)\%$ success rate.

\subsection{Environmental Change}

To test the ability of our model to adapt and learn continually, we simulate a mechanical change in the NICO robot by locking its shoulder motor in a fixed position after having trained with full-motion capabilities. The internal model is tested for $500$ iterations, repeated $5$ times, for the ability to reach the target after the shoulder was modified. The median Euclidean error from the gaze control model is $2.16(\pm 0.09)$ pixels. 
We then allow the model to adapt and train with the new kinematics. For the adapted model, the median Euclidean error is $5.08(\pm 1.76)$ cm, while the median Euclidean error for the original model is $6.93(\pm 1.57)$ cm. 
The minimum error for the adapted model is $1.61(\pm 3.26)$ cm, while the maximum error is $26.70(\pm 8.84)$ cm. For the original model, the minimum error 
is $1.93(\pm 3.87)$ cm, and the maximum error $26.83(\pm 8.24)$ cm. 
With the original model, the target is reached $206.9(\pm 6.64)$ times, which corresponds to a $20.68(\pm 0.66)\%$ success rate. With the adapted model, the target is successfully reached $230.8(\pm 11.17)$ times, which corresponds to a $23.08(\pm 1.11)\%$ success rate. 

While the overall decreased success in reaching can be mainly attributed to the locked shoulder joint that makes it impossible to reach certain poses, the trained model reaches the target $23.9$ times more on average than the untrained model. This result shows that our model successfully adapts to severe changes in the robot's mechanics and continuously learns from interaction with the environment.

\section{Conclusion}

In this paper, we presented and investigated an approach that enables the autonomous learning of object reaching with the NICO humanoid robot. The necessary sensorimotor information was acquired purely through 
body-environment interactions. In contrast to supervised learning approaches, the reach action is addressed and learned indirectly via a developmental approach by first learning how to control the gaze to situate NICO in the environment, then learning how to control the arm, and then finally learning eye-hand coordination. This enables NICO to associate hand positions with a direction of gaze, of which the inverse task is the reach action. Grow-When-Required models were trained on and capture the temporal co-occurence of the temporal sensorimotor information through direct connections between the networks via Hebbian learning. The approach of modeling the internal model with GWR networks allows each network to add neurons whenever the sensorimotor information cannot be sufficiently matched by any neuron, which results in a continuous growth of the network.

The results on the task of reaching without any prior information indicate that the internal model with GWRs learns the unimodal information and captures the associations between the temporally correlated sensorimotor information by enabling NICO to reach a target with a median accuracy of $2.44(\pm 0.10)$ cm. The observed reach actions are relatively coarse, which is in line with findings from developmental studies where human infants try coarse reach actions with most of them failing, as the motoric competence and depth perception that would be required for precise tasks is not fully developed yet \cite{schillaci_sensorimotor_2014, law_saccades_2014, law_infant_2011}. Furthermore, we demonstrate that the internal model is adaptable to environmental or mechanical changes---an already trained model was able to successfully adapt to a locked shoulder joint.

In future work, we will implement overlapping neural structures \cite{earland_overlapping_2014} in the GWRs to improve the generalization ability, as well as dimensionality reduction techniques and pruning methods to lower the complexity of each network to improve performance and learning time. For the perception of depth, we extend our model to acquire feedback from the arm control model, where the sensorimotor information is used to predict the current distance between the eye and the hand. Furthermore, we will implement our proposed model into the physical NICO, which needs more sophisticated exploration strategies like goal babbling and/or reward-driven self-organization \cite{aswolinskiy_rm-sorn_2015} mechanisms to improve the dataset generation, and compare the reach performance and the adaptability to existing solutions.

\end{document}